\documentclass[11pt]{article}
\usepackage{fullpage}

\usepackage{lipsum}
\usepackage{graphicx} 
\usepackage{subfig}
\usepackage{natbib}

\usepackage{wrapfig}

\usepackage{algorithm}
\usepackage{algorithmic}

\usepackage{epsfig}
\usepackage{amsthm}
\usepackage{amsmath}
\usepackage{amssymb}
\usepackage{color}
\usepackage{epstopdf}

\newcommand{\argmin}{\operatornamewithlimits{argmin}}

\let\oldbibliography\thebibliography
\renewcommand{\thebibliography}[1]{\oldbibliography{#1}
\setlength{\itemsep}{2pt}}

\usepackage{times}

\usepackage{enumitem}

\usepackage{hyperref}

\usepackage{epsfig}
\usepackage{amsthm}
\usepackage{amsmath}
\usepackage{amssymb}
\usepackage{color}
\usepackage{epstopdf}

\usepackage{hyperref}

\newcounter{thm_counter}

\newtheorem{theorem}[thm_counter]{Theorem}
\newtheorem{lemma}[thm_counter]{Lemma}

\def\abovestrut#1{\rule[0in]{0in}{#1}\ignorespaces}
\def\belowstrut#1{\rule[-#1]{0in}{#1}\ignorespaces}

\def\abovespace{\abovestrut{0.20in}}

\def\belowspace{\belowstrut{0.10in}}

\def\E{\mathbb{E}}

\def\P{\mathbb{P}}
\def\PB{\mathbb{P}_{\text{Ber}(p)}}
\def\PU{\mathbb{P}_{\text{Unif}(k)}}

\newcommand{\beq}{\begin{equation}}
\newcommand{\eeq}{\end{equation}}

\newcommand{\R}{\mathbb{R}}
\def\G{\mathcal{G}}
\def\1{{\bf{1}}}
\def\0{{\bf{0}}}
\def\I{{\bf{I}}}
\def\w{{\bf w}}
\def\u{{\bf u}}
\def\v{{\bf v}}

\def\x{{\bf x}}
\def\y{{\bf y}}

\def\a{{\bf a}}

\def\c{{\bf c}}
\def\d{{\bf d}}
\def\bfXi{{\boldsymbol \xi}}

\def\locp{\mbox{\sc Fista-locp}}
\def\licp{\mbox{\sc Fista-licp}}
\def\pcp{\mbox{\sc Fista-pcp}}
\def\admm{\mbox{\sc Admm}}

\def\fista{\mbox{\sc Fista}}

\def\sjwcommentsolved#1{}
\def\jlcommentsolved#1{}

\title{Exclusive Sparsity Norm Minimization with Random Groups via Cone Projection}
  
\author{Yijun Huang and Ji Liu\\
\{huangyj0, ji.liu.uwisc\}@gmail.com\\
Department of Computer Sciences, University of Rochester}
\date{\today}

\begin{document}
\maketitle 

\begin{abstract}
Many practical applications such as gene expression analysis, multi-task learning, image recognition, signal processing, and medical data analysis pursue a sparse solution for the feature selection purpose and particularly favor the nonzeros \emph{evenly} distributed in different groups. The exclusive sparsity norm has been widely used to serve to this purpose. However, it still lacks systematical studies for exclusive sparsity norm optimization. This paper offers two main contributions from the optimization perspective: 1) We provide several efficient algorithms to solve exclusive sparsity norm minimization with either smooth loss or hinge loss (non-smooth loss). All algorithms achieve the optimal convergence rate $O(1/k^2)$ ($k$ is the iteration number). To the best of our knowledge, this is the first time to guarantee such convergence rate for the general exclusive sparsity norm minimization; 2) When the group information is unavailable to define the exclusive sparsity norm, we propose to use the random grouping scheme to construct groups and prove that if the number of groups is appropriately chosen, the nonzeros (true features) would be grouped in the ideal way with high probability. Empirical studies validate the efficiency of proposed algorithms, and the effectiveness of random grouping scheme on the proposed exclusive SVM formulation. 
\end{abstract}

\section{Introduction}

The exclusive sparsity norm ($\ell_e$ norm, also namely $\ell_{1,2}$ norm in some literatures) is defined as
\begin{equation} 
\|\w\|_{e} := \sqrt{\sum_{g\in \G} \|\w_{g}\|_1^2}
\end{equation}
where $\w\in \R^n$, $g_i~(i=\{1,2,\cdots, m\})$ is a subset of $\{1,2,\cdots, n\}$, and $\G=\{g_1,g_2,\cdots, g_m\}$ is the super set of groups. The groups $g_i$'s may have overlaps. The exclusive sparsity norm regularization serves to enforce a structural sparsity on the solution such that its nonzeros are (roughly) evenly distributed in different groups.

\begin{figure}[t]
\begin{center}
\vspace{-5mm}
\includegraphics[width=0.6\columnwidth]{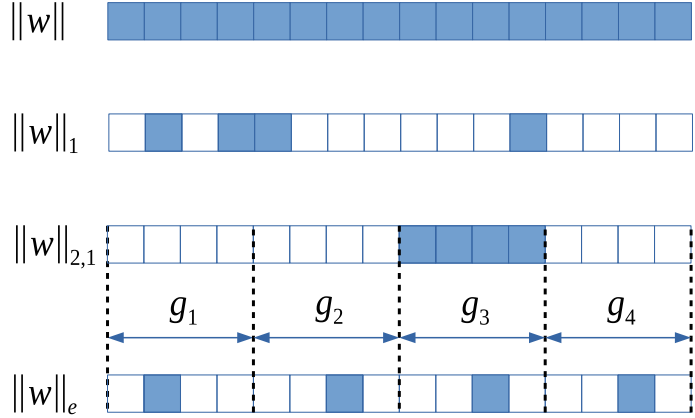}
\vspace{-4mm}
\caption{Sparsity structure induced by different norms: $\ell_2$ norm, $\ell_1$ norm, $\ell_{2,1}$ norm, and $\ell_e$ norm. Blue blocks indicate nonzeros and white blocks indicate zeros.}
\vspace{-4mm}
\label{fig:sparsity}
\end{center}
\end{figure}

Figure~\ref{fig:sparsity} shows an example to illustrate the difference of the sparsity pattern induced by the $\ell_e$ norm regularization from several other commonly used (sparse) regularizations including $\ell_2$ norm $\|\w\|$, $\ell_1$ norm $\|\w\|_1$, and $\ell_{2,1}$ (group sparsity) norm $\|\w\|_{2,1}$. We can observe that 
\begin{itemize}[noitemsep,nolistsep,leftmargin=*]
\item the $\ell_2$ norm does not enforce sparsity (actually usually leads to a dense solution); 
\item the $\ell_1$ norm enforces a sparse solution but has no any structural pattern; 
\item the $\ell_{2,1}$ norm enforces a group sparsity, i.e., nonzeros concentrate in the same group;
\item the $\ell_e$ norm provides a dense solution in the group level but a sparse solution inside of groups.
\end{itemize}
To further explain the difference among these norms above, we assume that there are two important features $f_i$ and $f_j$ in two disjoint groups. They are similar and highly correlated. $\|\w\|$ prefers to select both of them; $\|\w\|_1$ prefers to select the one which gives a lower objective function value; $\|\w\|_{2,1}$ prefers to select the one with more group members selected; and $\|\w\|_{e}$ prefers to select the one with fewer group members selected. 

Since the exclusive sparsity norm enforces a special and unique sparse structure, it has been widely used and received many successes in applications including gene expression analysis \citep{breheny2009penalized, kim2012tree}, siginal processing \citep{bayram2014primal, balazs2013adapted, kowalski2013social}, multi-task learning \citep{wang2014multiplicative, lozano2012multi}, image recognition / classification \citep{chen2011multi, liu2011multi} and medical data analysis \citep{lin2014integrative, vidyasagar2014machine}. 

While the $\ell_1$ norm, $\ell_2$ norm, and $\ell_{2,1}$ norm have received extensive studies, it still lacks systematical studies for the exclusive sparsity $\ell_e$ norm on twofold: optimization (how to efficiently solve $\ell_e$ norm minimization problem) and consistency analysis (under what condition the solution to the $\ell_e$ norm regularized formulation converges to the true model asymptotically?). This paper mainly considers the optimization aspect and leave the consistency analysis as the future work.

We mainly offer two contributions. First, this paper considers two types of $\ell_e$ norm minimization problems 1) smooth convex loss + $\|\cdot\|^2_e$ and 2) hinge loss (non-smooth convex function) + $\|\cdot\|^2_e$. To efficiently solve the $\ell_e$ norm minimization problems, the key step is to solve the proximal problem. Due to the difficulty of precisely solving the proximal step, it still lacks efficient algorithms to solve the general $\ell_e$ norm minimization problems, especially for the case where groups have overlaps. We propose several simple tricks to reformulate the original formulations, derive efficiently solvable proximal steps, and provide three optimization algorithms ($\pcp$, $\locp$, $\licp$) achieving the optimal convergence rate by adopting the $\fista$ framework. To the best of our knowledge, this is the first time to obtain such optimal convergence rate $O(1/k^2)$ for the general $\ell_e$ norm minimization problems. In addition, a new SVM (E-SVM) formulation is proposed.

Second, existing applications of the exclusive sparsity norm usually assume that groups have been well defined. However, if the group information is unavailable, how to define reasonable groups? This paper proposes to use the random grouping scheme to obtain groups and proves that each group will have roughly equal number of true features with high probability by using this scheme. Empirical studies validate the efficiency of the proposed algorithms and effectiveness of the proposed random grouping scheme.    


{\noindent \bf Notation} 
Throughout the whole paper, we use the following notations:
\begin{itemize}
\item Both $g$ and $g_i$ denote the subset of $\{1, 2, \cdots, n\}$; other plain small letters denote scalars, for example, $a, x, \alpha$. 
\item $n$ denotes the dimension of the feature vector. $m$ denotes the number of groups. 
\item Bold small letters denote vectors, for example, $\a, \u, \x, \w$. 
\item $\mathcal{G}$ is the superset of all groups $g_1,\cdots, g_m$; other capital letters denote matrices, for example, $A, X$. 
\item $|\cdot|$ returns a vector consisting of absolute values of components in ``$\cdot$'', if ``$\cdot$'' is a vector. $|\cdot|$ returns the cardinality of ``$\cdot$'', if ``$\cdot$'' is a set. 
\item $\w_g \in \R^{|g|}$ is the subvector of $\w$ consisting of elements of $\w$ in the index set $g$. $\w_i$ is the $i$th component of $\w$. $\1$ denotes a vector with all $1$'s. We will also use the notation $\w^g \in \R^{|g|}$, which is a free vector named by $\w^g$ and whose value has nothing to do with $\w$. 
\item $\|\w\|_2$ or $\|\w\|$ denotes the $\ell_2$ norm of $\w$: $\sqrt{\sum_i \w_i^2}$.
\item $\|\w\|_1$ denotes the $\ell_1$ norm: $\sum_i |\w_i|$.
\item $\|\w\|_\infty$ is the infinity norm: $\max_i |\w_i|$. 
\item ${\bf I}_{\text{condition}}(\x)$ is the indicator function which returns $0$ if $\x$ satisfies the ``condition''; otherwise $+\infty$.
\end{itemize}

{\noindent \bf Organization} 
The related work is reviewed in Section \ref{sec:rel}. In Section \ref{sec:opt}, we provide preliminary results first, then propose efficient algorithms for solving a general problem in the form ``smooth loss + exclusive sparsity regularization'', and in the last propose the exclusive SVM formulation ``hinge loss + exclusive sparsity regularization'' and the efficient algorithm to solve it. We discuss the random grouping scheme in Section \ref{sec:rand}. We report the empirical study for the proposed algorithms and formulations in Section \ref{sec:exp} and conclude this paper in Section \ref{sec:conclusion}. All proofs are provided in Supplemental Materials.

\section{Related Work} \label{sec:rel}
Structural sparsity induced by regularization terms have been widely used recently for the feature selection purpose \citep{bach2012structured, jenatton2011structured}. Both theoretical and empirical studies have suggested the powerfulness of structure sparsity for feature learning, e.g., LASSO \citep{tibshirani1996regression}, group LASSO \citep{yuan2006model}, exclusive LASSO \citep{zhou2010exclusive}, fused LASSO \citep{tibshirani2005sparsity, liu2013guaranteed}, and generalized LASSO \citep{roth2004generalized}. 
We refer readers to \citet{bach2012structured} for comprehensive review.

The $\ell_1$ norm minimization problem, particularly the LASSO formulation \citep{tibshirani1996regression}, has received extensive attentions. Many efficient algorithms have been proposed to solve it, including proximal gradient descent \citep{wright2009sparse}, accelerated gradient descent \citep{nesterov2004introductory}, $\fista$ \citep{beck2009fast}, coordinate descent \citep{wu2008coordinate}, accelerated coordinate descent \citep{peter2014}, and interior point method \citep{koh2007interior}. We refer readers to \citet{tropp2010computational} for comprehensive review of the $\ell_1$ norm optimization.

The $\ell_{2,1}$ (group sparsity) norm minimization, particularly the group LASSO (without overlaps) \citep{yuan2006model}, can be solved using similar optimization algorithms to $\ell_1$ norm problems by simply treating each group as an element \citep{liu2009multi}. For the the group LASSO with overlaps, one can use the $\admm$ framework to solve it \citep{boyd2011distributed}, where the key step can still apply those approaches for non-overlapping formulations.

The exclusive sparsity regularization was originally defined on matrices for the feature selection purpose in multi-task learning \citep{zhou2010exclusive}. A subgradient algorithm is proposed to solve this problem, but leads to a slow convergence rate $O(1/\sqrt{k})$, where $k$ is the number of iterations. \citet{KongNIPS14} considers to minimize an objective ``$\text{least squares loss}+\lambda \|\w\|^2_e$'' and proposes a reweighed approach to solve it. However, the proposed reweighted approach only guarantees convergence, which means that the convergent point is not necessarily the optimal solution. \citet{chen2011multi} considers to minimize an variant objective ``$\text{least squares loss}+\lambda \|D\w\|^2_e$''. They apply the Nesterov's smoothing scheme to smooth the regularization term and the Nesterov's accelerated approach to solve an approximate smooth objective function \citep{nesterov2005smooth}. This algorithm leads to a convergence rate $O(1/k)$. To the best of our knowledge, none of existing algorithms can strictly achieve the optimal convergence rate $O(1/k^2)$ for the general exclusive sparsity norm minimization even the loss function is smooth.    


\begin{algorithm}[H]                      
\caption{The General Framework of $\fista$}      
\label{alg:FISTA}                           
\begin{algorithmic}[1]                    
    \REQUIRE $\x^{\text{old}}$, $\gamma$ (steplength)
    \ENSURE $\x^{\text{new}}$
\STATE Initialize $t^{\text{old}} = 1$, $\bar{\x}=\x^{\text{old}}$;
\WHILE {not converge}
\STATE $\x^{\text{new}}=\text{Prox}_{\gamma H(\x)}(\bar{\x} - \gamma \nabla F(\bar{\x}))$; \label{alg:step:prox}
\STATE $t^{\text{new}} = {1\over 2} + {1\over 2}\sqrt{1+4 t^{\text{old}}}$;
\STATE $\bar{\x} = \x^{\text{new}} + (\x^{\text{new}} - \x^{\text{old}}) (t^{\text{old}} -1) / t^{\text{new}}$;
\STATE $\x^{\text{old}} = \x^{\text{new}}$, $t^{\text{old}} = t^{\text{new}}$;
\ENDWHILE
\end{algorithmic}
\end{algorithm}

\section{Optimization} \label{sec:opt}
We consider two types of exclusive sparsity norm minimization problems: smooth loss function and non-smooth loss (specifically hinge loss) function. Several efficient algorithms would be provided, with the guarantee of the optimal convergence rate $O(1/k^2)$.


\subsection{Preliminary Results} \label{sec:pre}
For optimization problems with the structure: ``smooth loss $F(\x)$ + non-smooth function $H(\x)$''
\begin{align}
\min_{\x}\quad F(\x) + H(\x), \label{eq:slnr}
\end{align}
the Nesterov type of accelerated algorithms \citep{nesterov2007gradient}, for example, $\fista$ \citep{beck2009fast} in Algorithm~\ref{alg:FISTA}, is often considered as one of the most efficient gradient based algorithms, since it achieves the optimal convergence rate $O(1/k^2)$ where $k$ is the number of iterations. $H(\x)$ could be the regularizer such as $\|\x\|_1$ and $\|\x\|$ or the indicator function representing the constraint, for example, ${\bf I}_{\x\geq 0}(\x)$. One can see the key step in $\fista$ is Step~\ref{alg:step:prox} (the proximal step) in Algorithm~\ref{alg:FISTA}, which essentially solves the following problem:
\begin{equation}
{\text{Prox}}_{\gamma H(\x)}(\c):=\argmin_{\x}\quad {1\over 2} \|\x - \c\|^2 + \gamma H(\x)
\label{eq:prox}
\end{equation}
where $\c = \bar{\x} - \gamma\nabla F(\bar{\x})$. The steplength $\gamma$ can be set as any positive constant smaller than $(\max_{x} \|\nabla^2 F(x)\|)^{-1}$ or decided by using the linear search scheme \citep{beck2009fast}. The difficulty of solving \eqref{eq:slnr} is decided by the complexity of $H(\x)$. When $H(\x)$ is simple enough such as $\|\x\|_1$, $\|\x\|$, or ${\bf I}_{\x\geq 0}(\x)$, the proximal step \eqref{eq:prox} is easy. However, if $H(\x)$ is $\|\x\|_e$, it is difficult to solve efficiently even all groups have no overlaps. Therefore we need reformulation work to simplify the proximal steps. The following preliminary results play the key roles in reformulation and solving proximal steps. 



\begin{algorithm} [htp!]                     
\caption{$[\x, y] = \text{P}_{1}^{\zeta}(\a, b)$}          
\label{alg:projl1cone}                           
\begin{algorithmic}[1]                    
    \REQUIRE $\a\in \R^d$, $b \in \R$, $\zeta \in \R$ 
    \ENSURE $\x\in \R^d$, $y\in \R$
\STATE Sort the sequence $\{|\a_j|~|~j=1,\cdots, d\}$ in the decreasing order and denote $|\a|^{(j)}$ the $j$th largest element of $\a$ in the absolute value sense and define $|\a|^{(d+1)}$ as $0$;
\STATE $t=0$;
 	\FOR{$j=1:d$}
	\STATE $t = t + |\a|^{(j)}$;
	\STATE $\delta=(t-b)/(\zeta^{-1}+j)$;
	\IF{$|\a|^{(j+1)}\leq \delta \leq |\a|^{(j)}$}
	\STATE $\x = \text{sgn}(\a)\odot \max(\0, |\a|-\delta)$;
\STATE $y = \|\x\|_1$;
\STATE Return;
	\ENDIF
\ENDFOR
\STATE $\x = \a$;
\STATE $y=b$;
\end{algorithmic}
\end{algorithm}

\begin{algorithm} [htp!]                     
\caption{$[\x, y] = \text{P}_{\infty}^{\zeta}(\a, b)$}          
\label{alg:projection_inf}                           
\begin{algorithmic}[1]                    
    \REQUIRE $\a\in \R^d$, $b \in \R$, $\zeta \in \R$ 
    \ENSURE $\x\in \R^d$, $y\in \R$
\STATE $y=0$;
\STATE $\x={\bf 0}$;
\STATE Sort the sequence $\{|\a_j|~|~j=1,\cdots, d\}$ in the increasing order and denote $|\a|^{(j)}$ the $j$th smallest element of $\a$ in the absolute value sense and define $|\a|^{(0)}$ as $0$;
\IF{$b \geq |\a|^{(d)}$}
\STATE $y = b$;
\STATE $\x = \a$;
\STATE Return;
\ENDIF
\STATE $t = 0$;
 	\FOR{$j=d-1:-1:0$}
	\STATE $t = t + |\a|^{(j+1)}$;
	\STATE $y = {\zeta b + t \over \zeta + d - j}$;
	\IF{$y \in (|\a|^{(j)},~|\a|^{(j+1)}]$}
\STATE $\x = \text{sgn}(\a) \odot \min(|\a| - y, 0)$;
\STATE Return;
\ENDIF
\ENDFOR
\end{algorithmic}
\end{algorithm}

We first propose an alternative way to represent $\|\w\|_e^2$ in Lemma~\ref{lem:dual}:
\begin{lemma} \label{lem:dual}
Let $\v^{g} \in \R^{|g|}$ denote the $|g|-$ dimensional vector. For any $\beta>0$, we have
\begin{align}
\|\w\|_e^2 = \sum_{g\in \mathcal{G}}\left( \max_{\v^g \in \R^{|g|}} \frac{2}{\beta}\langle \w_g, \v^g \rangle - {1\over \beta^2} \|\v^g\|_{\infty}^2\right).
\label{eq:lem:proj:l1}
\end{align}
\end{lemma}
This lemma is used to decompose the overlaps among groups in the dual space. It will be used to reformulate E-SVM. 
%

%
\begin{lemma} \label{lem:projl1cone}
Algorithm~\ref{alg:projl1cone} exactly solves ${\bf P}^{\zeta}_{1}(\a, b)$ -- the projection of $[\a; b]$ onto the $\ell_1$ norm cone $\{[\x; y]~|~\|\x\|_1 \leq y\}$:
\begin{equation}
\begin{aligned}
{\bf P}^{\zeta}_{1}(\a, b) := \argmin_{\x \in \R^d, y}\quad & {1\over 2} \|\x- \a\|^2 + {\zeta \over 2}(y-b)^2\quad \\
\text{s.t.}\quad & \|\x\|_1 \leq y.
\end{aligned}
\label{eq:proj1cone}
\end{equation}
\end{lemma}
Problem \eqref{eq:proj1cone} can be considered as a general version of the projection onto the $\ell_1$ cone. Although \eqref{eq:proj1cone} does not have the closed form, but its solution can be obtained from a search routine in Algorithm~\ref{alg:projl1cone}. Algorithm~\ref{alg:projl1cone} essentially gives a method to search a feasible point satisfying the KKT condition. The complexity of this algorithm is $O(d\log d)$. The key motivation behind this algorithm is the monotonicity of the optimal solution $\x^*$, that is, if $|\a_i| \geq |\a_j|$, then $|\x^*_i| \geq |\x^*_j|$. This lemma will be used in solving the special case (groups have no overlaps) of the first type problem.

Lemma \ref{lem:projection_inf} provides a solution to a similar problem to Lemma~\ref{lem:projl1cone}. The only difference lies on that the $\ell_1$ norm in the constraint is replaced by the $\ell_\infty$ norm. 
\begin{lemma}\label{lem:projection_inf}
Algorithm~\ref{alg:projection_inf} exactly solves ${\bf P}^{\zeta}_{\infty}(\a, b)$ -- the projection of $[\a; b]$ onto the $\ell_\infty$ norm cone $\{[\x; y]~|~\|\x\|_\infty \leq y\}$:
\begin{equation}
\begin{aligned}
{\bf P}_{\infty}^{\zeta}(\a, b) := \argmin_{\x \in \R^d,y} \quad & {1\over 2} \|\x-\a\|^2 + {\zeta\over 2}(y-b)^2 \\
\text{s.t.} \quad & \|\x\|_\infty \leq y
\end{aligned}
\label{eq:lem_proj}
\end{equation}
\end{lemma}
Problem \eqref{eq:lem_proj} can be considered as a general version of the projection onto the $\ell_\infty$ cone. 
The complexity of Algorithm~\ref{alg:projection_inf} is also $O(d\log d)$. This algorithm also follows the sprit of the monotonicity of the optimal solution. This lemma would be used in solving the problem with non-smooth loss function.

\subsection{Smooth Loss Function $+$ $\ell_e$ Regularization: $\pcp$ and $\locp$ Algorithms} \label{sec:smooth}
We consider the following general formulation
\begin{align}
\min_{\w}\quad f(\w) + {\lambda \over 2} \|\w\|^2_e.
\label{eq:pro_le2}
\end{align}
Note that here we do not assume groups $g$'s to be disjoint. One might ask why do not use $\phi \|\w\|_e$ as the regularizer. The reason lies on that ${\lambda \over 2} \|\w\|^2_e$ is easier to solve efficiently and leads to the same solution as using $\phi \|\w\|_e$ if $\lambda(\phi)$ is appropriately chosen. Note that $\|\w\|^2_e$ is still non-smooth. 

To directly apply the $\fista$ framework, one has to solve the proximal step \eqref{eq:prox} with $H(\cdot) = {\lambda \over 2} \|\w\|^2_e$. However, it is very difficult to solve it efficiently in general. Existing approaches apply iterative algorithms to \emph{approximately} solve this proximal step, for example, in \citet{KongNIPS14, yuan2011finite}, thus requiring heavy computation load (computing the inverse of a $n\times n$ matrix) and unable to theoretically ensure the convergence (rate). 

We propose to use a simple substitution to reformulate the problem \eqref{eq:pro_le2}, which largely simplifies the original formulation. The key idea is to decompose the variable $\w$ into two parts: positive part $\w_+ \geq \0$ and negative part $\w_- \geq \0$ such that $\w = \w_+ - \w_-$ and $\|\w_g\|_1 = \1^{\top} ((\w_+)_g + (\w_-)_g)$. It leads to an equivalent formulation to \eqref{eq:pro_le2}:
\begin{align}
\nonumber & \min_{\w_+\geq 0, \w_- \geq 0} ~ f(\w_+ - \w_-) + {\lambda \over 2} \sum_{g \in \mathcal{G}} (\1^{\top}(\w_+)_g+  \1^\top (\w_-)_g)^2
\nonumber \\ =&  \min_{\w_+, \w_- } \underbrace{f(\w_+ - \w_-) + {\lambda \over 2}(\w_+ + \w_-)^\top Q (\w_+ + \w_-)}_{=:F([\w_+; \w_-])} +
\nonumber \\ & \quad\quad \quad \underbrace{{\bf I}_{\w_+\geq 0}(\w_+) + {\bf I}_{\w_-\geq 0}(\w_-)}_{=: H([\w_+; \w_-])},
\label{eq:smooth_overlap}
\end{align}
where $Q$ is a positive semidefinite matrix and its elements are defined by 
\begin{align*}
Q_{ij} = 
\begin{cases}
  |\{g\in \mathcal{G}~|~\{i,j\}\subset g\}|,& \text{if } i \neq j\\
  |\{g\in \mathcal{G}~|~i\in g\}|,              & \text{otherwise.}
\end{cases}
\end{align*}
This simple trick turns the original non-smooth minimization \eqref{eq:pro_le2} into a smooth minimization with simple constraints. In other words, \eqref{eq:smooth_overlap} is an equivalent form of \eqref{eq:pro_le2}. \eqref{eq:smooth_overlap} can be easily solved by accelerated Nesterov approach or $\fista$, since the key proximal step has closed form
\begin{equation*}
\begin{aligned}
\text{Prox}_{\gamma H(\cdot)}([\c_+; \c_-]) = ~&\argmin_{\w_+, \w_- }~ {1\over 2}\|[\w_+; \w_-] - [\c_+; \c_-]\|^2 + \\ & \quad\quad\quad\quad H([\w_+; \w_-]) \\ 
=~&[\max(\0, \c_+); \max(\0, \c_-)].
\end{aligned}
\end{equation*}
Now we can apply the proximal operator to the $\fista$ framework in Algorithm~\ref{alg:FISTA}. Since the key step is the projection onto the positive cone, we call it \underline{p}ositive \underline{c}one \underline{p}rojection ($\pcp$) algorithm to distinguish the later algorithms. The main workload per iteration is just to compute the gradient, which is the minimal computation load for the gradient based approaches. To the best of our knowledge, this is the first approach to solve overlapped exclusive sparsity regularization with the optimal convergence rate $O(1/k^2)$. It is worth to point out that for the more general exclusive sparsity norm minimization ``$\text{smooth loss}+\sum_{g\in \mathcal{G}}\|\w_g\|^q_q$'' ($q\ge 1$), one can also apply this positive-negative decomposition trick to obtain the optimal convergence rate. 


Actually, if groups $g$'s are disjoint (or have no overlaps), we can derive a slightly more efficient algorithm, because the following proximal step can be solved efficiently (letting $H(\w) = {\lambda\over 2} \|\w\|^2_e$) 
\begin{align}
{\text{Prox}_{\gamma H(\cdot)}}(\c):=\argmin_{\w}\quad {1\over 2} \|\w - \c\|^2 + {\gamma\lambda \over 2} \|\w\|_e^2.
\label{eq:proximal_nooverlap}
\end{align}
The following will show that this problem can be solved efficiently due to Lemma~\ref{lem:projl1cone}. (If we use $\|\w\|_e$ as the regularizer, then \eqref{eq:proximal_nooverlap} is difficult to solve efficiently.) This why we are interested in \eqref{eq:pro_le2}. 

Because groups are disjoint, the proximal step can be split into a few subproblems
\[
\min_{\w_{g}} \quad {1\over 2} \|\w_g - \c_g\|^2 + {\gamma\lambda \over 2} \|\w_g\|_1^2\quad \forall g \in \mathcal{G}.
\]
To solve this problem, we can reformulated it into the form of \eqref{eq:proj1cone}
\begin{align*}
\min_{\w_g, t}\quad & {1\over 2} \|\w_g - \c_g\|^2 + {\gamma\lambda \over 2} t^2 \quad 
\\
\text{s.t.}\quad &\|\w_g\|_1 \leq t
\end{align*}
whose solution is exactly provided by ${\bf P}_1^{\gamma\lambda}(\c_g, 0)$.

Now we can simply apply the proximal operator defined in \eqref{eq:proximal_nooverlap} to the $\fista$ framework to obtain an algorithm with the optimal convergence rate. Since the proximal step in this algorithm is the projection onto the $\ell_1$ cone, we call this algorithm as \underline{L}-\underline{o}ne \underline{c}one \underline{p}rojection ($\locp$) algorithm. One can verify that the computation load per iteration is still on the gradient, that is, $O(n^2)$, if the number of samples is proportional to $n$.

\subsection{Exclusive SVM (Non-smooth Loss + $\ell_e$ Regularization): $\licp$ Algorithm} \label{sec:non-smooth}
This section considers the case where the loss function is non-smooth. In general, it is difficult to find an efficient algorithm to minimize non-smooth objective function. To solve such general problems, one can apply the subgradient algorithm, which leads to a convergence rate $O(1/\sqrt{k})$. We only consider a specific non-smooth loss function ``hinge loss'' which is the most popular non-smooth loss function in machine learning and data mining due to SVM. In particular, we are interested in solving the following exclusive SVM (E-SVM) formulation:
\begin{equation}
\begin{aligned}
\min_\w \quad \sum_i & \max(1-Z_i^{\top}\w, 0) + {\alpha \over 2}\|\w\|^2 + {\beta \over 2} \|\w\|_e^2,
\end{aligned}
\label{eq:ESVM}
\end{equation}
where groups $g_i$'s defined in the $\ell_e$ norm may have overlaps. This formulation is motivated by finding a linear classifier only defined on a few features which evenly distributed in different groups. ``Hinge loss + exclusive sparsity regularization'' in \eqref{eq:ESVM} is a very natural idea, but people usually try to avoid solving it mainly due to the lack of efficient algorithms. For example, to avoid solving \eqref{eq:ESVM}, \citet[see Section 5.2]{KongNIPS14} pretend the classification problem to be the regression problem by using the exclusive LASSO formulation for the feature selection purpose. The section derives an efficient algorithm to solve \eqref{eq:ESVM} with the optimal convergence rate $O(1/k^2)$. 

Apparently, it is ineligible to apply the $\fista$ framework directly, due to the non-smooth ``hinge loss'' and the overlapped exclusive sparsity regularizer $\|\w\|^2_e$. The following reformulation uses Lemma~\ref{lem:dual}  and strong duality to reformulate \eqref{eq:ESVM}:
\begin{theorem} \label{thm:esvm-re}
Let $[\u_*; \{\v^g_*\}_{g\in \mathcal{G}}]$ be the solution of the following problem:
\begin{align}
\nonumber [\u_*; \{\v^g_*\}_{g\in \mathcal{G}}]=\argmin_{\u, \{\v^g\}_{g\in \mathcal{G}}}&  \underbrace{{1\over 2\alpha} \left\|Z \u - \sum_{g\in \mathcal{G}}\overline{\v^g}\right\|^2 - \1^{\top} \u }_{=:F([\u; \{\v^g\}_{g\in \mathcal{G}}])} + 
\\ & \underbrace{{1 \over 2\beta}\sum_{g\in \mathcal{G}}\|\v^g\|_\infty^2 + \I_{\0\leq \u \leq \1}(\u)}_{=:H([\u; \{\v^g\}_{g\in \mathcal{G}}])},
\label{eq:ESVM_dual_max}
\end{align}
where $Z_i$ (the $i$th column of matrix $Z$) denotes $y_iX_i$ and $\overline{\v^g} \in \R^n$ to denote the extended version of $\v^g \in \R^{|g|}$, that is, the elements in $g$ of $\overline{\v^g}$ take the corresponding values of $\v^g$ and zeros in the rest. Then we have that the solution $\w_*$ to E-SVM in \eqref{eq:ESVM} can be obtained by 
\[
\w_*={\alpha^{-1}}\left(Z \u_* - \sum_{g\in \mathcal{G}}\overline{\v^g}_*\right).
\]
\end{theorem}

\eqref{eq:ESVM_dual_max} essentially defines a dual formulation of \eqref{eq:ESVM}. We can reconstruct the solution to \eqref{eq:ESVM} by solving \eqref{eq:ESVM_dual_max}, which actually fits the structure ``smooth function $F(\cdot)$ + non-smooth function $H(\cdot)$'' in \eqref{eq:slnr}. To apply the $\fista$ framework in Algorithm~\ref{alg:FISTA}, we only need to show an efficient algorithm to compute the proximal step defined in \eqref{alg:step:prox}:
\begin{align}
\nonumber
&\text{Prox}_{\gamma H(\cdot)}([\c; \{\d^g\}_{g\in \mathcal{G}}]) \\
\nonumber
:=&\argmin_{\u, \v^g:~g\in\mathcal{G}} {1\over 2}\|\u-\c\|^2 + \gamma {\bf I}_{\u\in [\0,\1]}(\u) + \\ & \quad\quad\quad\quad \sum_{g\in \mathcal{G}}\left({1\over 2} \|\v^g - \d^g\|^2 + {\gamma\over 2\beta} \|\v^g\|^2_\infty\right).\label{eq:proxinf}
\end{align}

Since $H(\u, \{\v^g:~g\in \mathcal{G}\})$ is separable in terms of all variables $\u$ and $\v^g (g\in \mathcal{G})$. $\u$ can be simply solved by projecting $\c$ to the feasible region $[\0, \1]$. All $\v^g$'s are computed solving the following problem:   
\[
\min_{\v^g}\quad {1\over 2} \|\v^g - \d^g \|^2 + {\gamma\over 2\beta} \|\v^g\|^2_{\infty}\quad \forall g\in \mathcal{G}.
\]
To solve each subproblem, we can reformulate it into the form of \eqref{eq:lem_proj}
\begin{align*}
\min_{\v^g}\quad & {1\over 2} \|\v^g- \d^g\|^{2} + {\gamma \over 2\beta} t^2 \quad\quad 
\\ \text{s.t.}\quad &  \|\v^g\|_\infty \leq t.
\end{align*}
From Lemma~\ref{lem:projection_inf}, each subproblem can be solved by ${\bf P}_{\infty}^{\gamma/\beta}(\d^g, 0)$ in Algorithm~\ref{alg:projection_inf}. 
Since the key proximal step mainly uses the projection onto the infinity cone, we call this algorithm as L-infinity cone projection ($\licp$) algorithm.

\section{Random Grouping Scheme} \label{sec:rand}


The group information needs to be known to define the exclusive sparsity norm. In some cases such as multi-task feature selection \citep{zhou2010exclusive} and image classification \citep{chen2011multi}, groups have been defined in a natural manner, for example, features are grouped in terms of kinds. However, sometimes the natural group information is unavailable or does not make sense for some specific purpose. Therefore, the key question is how to construct groups for defining the exclusive sparsity regularization. 

The exclusive sparsity regularization tends to provide a solution with nonzeros (selected features) evenly distributed in different groups. The ideal grouping result is that all $s$ true (really important) features are evenly distributed in different groups (of course we have no idea what features are true or important beforehand). This motivates us to use the random grouping scheme, that is, uniformly randomly split all features indexed by $\{1,2,\cdots, n\}$ into $m$ disjoint groups with roughly equal size. Apparently, the ideal case is that each group is assigned $s/m$ true features. Although the random grouping scheme does not necessarily achieve the ideal case, it still can provide a reasonable grouping result. Theorem~\ref{thm:even} basically shows that the number of true features assigned to any group would be roughly equal.



\begin{theorem}\label{thm:even}
Assume that we have $s$ true features among $n$ features, that is, the true model only has $s$ nonzero elements ($s\leq n$). The feature index set $\{1, 2,\cdots, n\}$ is uniformly randomly split into $m$ groups with roughly equal size.\footnote{The size difference between two groups is $1$ maximally.} If the chosen number of groups $m$ satisfies $s\geq O(t^{-2}m\log m)$, where $t$ could be any number in the range $(0,1)$, then with high probability,\footnote{``With high probability'' means that the probability converges to $1$ when the number of true features $s$ converges to infinity.} the following holds
\[
\frac{\max_j |\Omega_j|}{\min_j |\Omega_j|} \leq \frac{1+t} {1-t}.
\] 
where $\Omega_j \subset g_j$ denote the index set of true features assigned to group $j\in \{1,2,\cdots, m\}$.
\end{theorem}
Note that Theorem~\ref{thm:even} is independent to the total number of features $n$. The only requirement in this theorem is that the number of groups $m$ should be set properly to satisfy $s\geq O(t^{-2}m \log m)$. One can trivially set $m=1$ to meet this condition, it reduces the $\ell_e$ norm to the plain $\ell_1$ norm, which does not take any benefits from the $\ell_e$ norm. Therefore, intuitively one always expect to set $m$ to a value as large as possible. A simple way to choose $m$ to satisfy the condition is $m=\text{constant}\times s/\log s$. 







\section{Experiments} \label{sec:exp}

This section mainly validates the efficiency and the effectiveness of proposed algorithms, formulations, and the random grouping scheme. 

\begin{figure}[H]
\vspace{-3mm}
\centering
\subfloat[][Efficiency comparison on non-overlapped E-LASSO among Kong approach, PLS-DN, LELR, and the proposed algorithms ($\pcp$ and $\locp$).]{\includegraphics[width=0.48\textwidth]{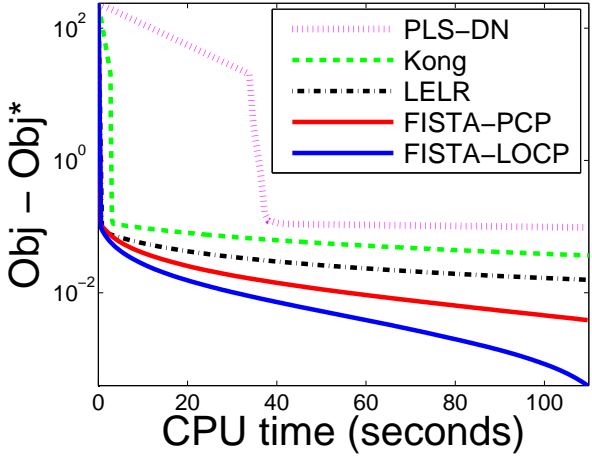}\label{fig:LOCP:synthetic_result}} \quad
\subfloat[][Efficiency comparison on overlapped E-LASSO among Kong approach, PLS-DN, LELR, and the proposed algorithm $\pcp$.]{\includegraphics[width=0.48\textwidth]{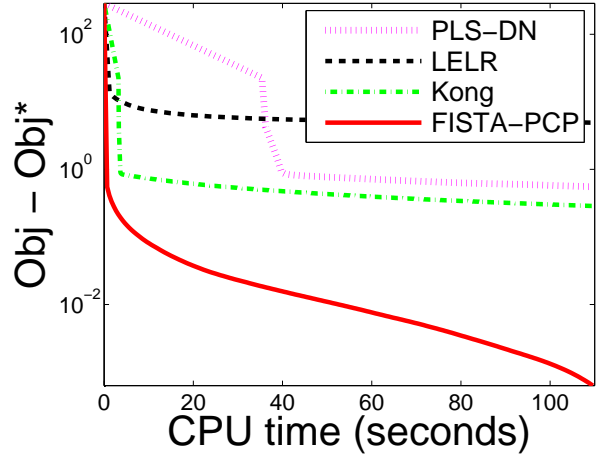}\label{fig:LOCP_OL:synthetic_result}} \\ 
\subfloat[][Efficiency comparison on overlapped E-SVM between $\admm$ and $\licp$.]{\includegraphics[width=0.48\textwidth]{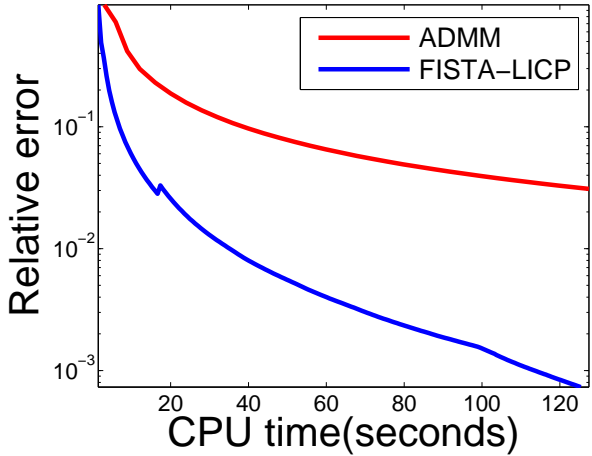}\label{fig:ADMM_FISTA:synthetic_result}} \quad
\subfloat[][Accuracy comparison on synthetic data among SVM, S-SVM, and the proposed formation E-SVM / E-SVM(R).]{\includegraphics[width=0.48\textwidth]{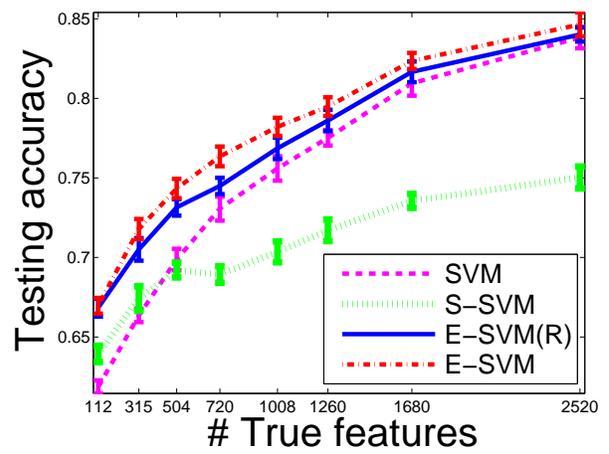}\label{fig:ESVM:synthetic_result}}
\caption{Plots for Experiments}
\vspace{-3mm}
\label{fig:exp}
\end{figure}

\subsection{Efficiency} \label{sec:exp:elasso}
This section conducts numerical simulation to validate the efficiency of proposed algorithms: $\pcp$, $\locp$, and $\licp$. 

For $\pcp$ and $\locp$, we consider the most popular exclusive LASSO (E-LASSO) formulation \citep{zhou2010exclusive}: ${1\over 2}\|X\w - \y\|^2 + \lambda \|\w\|^2_e$.


We compare $\pcp$ and $\locp$ to several recent algorithms of solving E-LASSO, including PLS-DN \citep{yuan2011finite}, Kong's approach \citep{KongNIPS14}, and LELR \citep{chen2011multi}. All algorithms are implemented in Matlab. We try our best to tune optimization parameters in all algorithms and report their best performance.



We first consider the nonoverlapped case. The elements in data matrix $X\in \R^{400 \times 4000}$ are generated from i.i.d. Gaussian distribution $\mathcal{N}(0, 1)$. The true model $\w^*\in \R^{4000}$ has been split into $100$ disjoint groups with $4$ nonzeros in each group. All nonzeros are generated from i.i.d. $\mathcal{N}(0,1)$. The observation vector $\y$ is generated from $\y=X \w^* + 0.01\times \mathcal{N}(0, 1)$. $\lambda$ is set as $0.4/ \|\w^*\|_1)$.  We report the objective function value against the CPU time in Figure~\ref{fig:LOCP:synthetic_result}. We observe that $\locp$ is the most efficient and $\pcp$ is the second best. This observation is not surprising, because $\locp$ and $\pcp$ strictly guarantees the optimal convergence rate with light computation complexity per iteration.  

We then consider the overlapped case. The synthetic data is generated in the same manner as above except that $100$ groups in $\w^*$ are randomly generated and the group size has changed from $40$ to $140$ (groups are highly overlapped). Note that the $\locp$ algorithm is not available in this case anymore. We report the efficiency comparison of the rest algorithms in Figure~\ref{fig:LOCP_OL:synthetic_result}. We can observe that $\pcp$ algorithm outperforms other algorithms.


 

To see the efficiency of $\licp$, we compare it to the general optimization algorithm $\admm$ \citep{boyd2011distributed} (to the best of our knowledge, there is no specific optimization algorithm for solving E-SVM). We compare two algorithms on synthetic data with 5040 features and 720 samples. The details on how to generate the synthetic data is introduced in Section \ref{sec:exp:esvm}. We plot curves of the relative error ($\|\w^{k+1} - \w^k\| / \|\w^k\|$) against the running time for both algorithms in Figure~\ref{fig:ADMM_FISTA:synthetic_result} and observe that $\licp$ is more efficient than $\admm$. This observation is not surprising since the convergence rate for $\licp$ $O(1/k^2)$ is superior to the rate for $\admm$ $O(1/k)$ \citep{he20121}.  


\subsection{Effectiveness} \label{sec:exp:esvm}

The effectiveness of the exclusive sparsity regularization for smooth loss such as exclusive LASSO can be found in many existing literatures. We refer readers to \citep{zhou2010exclusive, kowalski2009sparsity, chen2011multi, KongNIPS14}. This section only validates the effectiveness of the exclusive sparsity regularization in non-smooth loss as well as the random grouping scheme through solving the E-SVM formulation~\eqref{eq:ESVM}. We consider two setups of E-SVM: 1) E-SVM -- group information is available and 2) E-SVM(R) -- group information is unavailable using the proposed random grouping scheme. $\licp$ is used to solve both of them. We compare E-SVM and E-SVM(R) to the standard SVM, sparse SVM (S-SVM) \citep{bi2003dimensionality}, and Kong's approach  \citep{KongNIPS14} on both synthetic data and real data.

The synthetic data is generated as follows. The true classifier $\w^* \in \R^n$ vector is randomly split into $m$ disjoint groups $g_1,g_2,\cdots, g_m \subset\{1, 2, \cdots, n\}$ with equal size. Each group $\w_{g_i}$ only has a single nonzero element, whose value is generated from Gaussian distribution $\mathcal{N}(0, 1)$. Then we generate the data matrix $X = [X_{+}, X_{-}] \in \R^{n \times m}$, where $X_+$ and $X_-$ denote $m/2$ positive class samples and $m/2$ negative class samples. (Note that the number of samples is equal to the number of groups $m$.) We first generate a matrix $X_0 \in \R^{n\times {m\over 2}}$ with elements generated from i.i.d. Gaussian $\mathcal{N}(0, 1)$. $X_{+}$ is obtained by $X_+(:,i) = X_0(:, i) + d\w^*$ for any positive sample $i$ and $X_{-}$ is obtained by $X_-(:,j) = X_0(:, j) - d\w^*$ for any negative sample $j$. Here, $d$ is used to control the accuracy. We tune the value of $d$ such that the misclassification rate using the true model $\w^*$ is around $10\%$. The testing data set is generated in the same manner and has the equal size. 

For the synthetic data, we fix the feature dimension as $5040$ and choose different values for the number of samples (groups) $m= 112, 315, 504, 720, 1008, 1260, 1680, 2520$. The group number in E-SVM(R) is set as $m/\log m$. All experiments using four variants of SVM formulation are repeated for $10$ times. We report the average accuracy of testing data for four algorithms in Figure~\ref{fig:ESVM:synthetic_result}. We observe that 1) the E-SVM outperforms other three algorithms, which is not surprising because it employs the structural sparsity information; 2) The E-SVM(R) is slightly worse than E-SVM, but still superior to SVM and S-SVM, which validates the effectiveness of the random grouping scheme; and 3) The performance of SVM, E-SVM and E-SVM (R) are comparable when the sparsity ratio is high.  

Next we compare the proposed E-SVM (R) to SVM, S-SVM, and Kong's approach \citep{KongNIPS14} (they use exclusive LASSO to select features and then apply SVM to only selected features) on real datasets. Since the group information is unavailable, we only use E-SVM(R) for comparison. Table \ref{real_result}~
reports the accuracy of four algorithms using cross validation. 
We observe that the proposed E-SVM(R) is overall superior to SVM, S-SVM, and Kong's approach. Note that the main purpose of this comparison is to validate the random grouping scheme and shows that E-SVM can be superior on some common real data sets under certain scenarios, rather than arguing that E-SVM can substitute SVM and S-SVM.

\vspace{6mm}
\begin{table*}[ht]
\caption{Accuracy on real datasets using SVM, S-SVM, Kong's approach, and E-SVM(R). Real datasets we use include Computer (ISOLET \citep{Bache+Lichman:2013}, Handwritten Digits (PCMAT \citep{zhao2010advancing}), Linguistic (PCMAT \citep{zhao2010advancing} and TDT2 \citep{TDT2_DCai}), Cancer (LEU \citep{golub1999molecular}, ALLAML \citep{15044459}, Colorectal \citep{alexandrov2009biomarker}, ProstateCancer (`JNCI 7-3-02') \citep{petricoin2002serum} and Prostate-GE \citep{Singh2002203}) and Social Media (TwitterHealthData \citep{sadilek2013nemesis}).}
\label{real_result}
\begin{center}
\begin{tabular}{l|c|cc|cccc}
\hline
\abovespace\belowspace
Data              & \#Feature & \#Samples & \#Flods  & SVM & S-SVM  & Kong  & E-SVM(R)  \\ 
\hline
\abovespace
ISOLET    &617    &1560    &41    &69.59\%    &67.37\%    &69.64\%    &{\bf 69.85}\% \\ 

MNIST    &784    &3119    &6    &95.92\%    &97.90\%    &97.69\%    &{\bf 98.11}\% \\ 

PCMAT    &3289    &1943    &31    &64.43\%    &55.05\%    &64.26\%    &{\bf 66.21}\% \\ 

TDT2    &36771    &885    &5    &53.84\%    &92.97\%    &79.77\%    &{\bf 94.07}\% \\ 

LEU    &3571    &72    &5    &83.66\%    &76.02\%    &83.67\%    &{\bf 84.71}\% \\ 

ALLAML    &7129    &72    &9    &74.48\%    &61.81\%    &74.31\%    &{\bf 74.65}\% \\ 

Colorectal    &16331    &112    &20    &80.17\%    &63.59\%    &{\bf 81.40}\%    &81.39\% \\ 

ProstateCancer    &15154    &89    &21    &{\bf 70.90}\%    &55.79\%    &70.89\%    &70.33\% \\ 

Prostate-GE    &5966    &102    &21    &64.46\%    &54.62\%    &67.45\%    &{\bf 67.65}\% \\ 

\belowspace
TwitterHealthData    &5087    &6873    &2    &71.96\%    &80.77\%    &81.17\%    &{\bf 82.67}\% \\ 

\hline
\end{tabular}
\end{center}
\end{table*}
\vspace{-2mm}

\section{Conclusion} \label{sec:conclusion}
This paper investigates the exclusive sparsity norm minimization problem. Several efficient algorithms are proposed to solve two types of objectives ``smooth loss + $\|\cdot\|^2_e$'' and ``hinge loss (non-smooth) + $\|\cdot\|^2_e$''. The optimal convergence rate $O(1/k^2)$ is guaranteed for all proposed algorithms. When the group information is not available, a random grouping scheme is proposed to define groups. This grouping scheme is proven to be capable of (roughly) evenly assigning true nonzeros to every groups with high probability. Empirical studies validate the efficiency of the proposed algorithms ($\pcp$, $\locp$, and $\licp$) and the effectiveness of the proposed E-SVM formulation and the random grouping scheme.   

\section*{Acknowledgments}

The authors would like to thank Dr. Deguang Kong at Samsung Research for sharing data sets and help with experiments. The authors would also thank Professor Henry Kautz and Roya Feizi at University of Rochester for sharing the TwitterHealthData used in Table~\ref{real_result}.






{
{\bibliography{reference}
\bibliographystyle{abbrvnat}
}
}

\newpage
\section*{Supplemental Materials: Proofs}



\noindent{\bf Proof to Lemma~\ref{lem:dual}}
\begin{proof}
We have
\begin{align*}
& \max_{\v^g \in \R^{|g|}}~\langle \w_g, \v^g \rangle - {1\over 2\beta} \|\v^g\|^2_\infty 
\\ = &
\max_t \max_{\|\v^g\|_\infty \leq t }~\langle \w_g, \v^g \rangle - {1\over 2\beta} t^2 
\\ = &
\max_t ~t\|\w_g\|_1 - {1\over 2\beta} t^2\quad \\ & (\text{from the fact that $\ell_1$ norm and $\ell_\infty$ norm are dual to each other.})
\\ = &
{\beta \over 2} \|\w_g\|_1^2.
\end{align*}
It completes the proof by summarizing over all $g\in \mathcal{G}$.
\end{proof}

Lemma \ref{lem:projl1cone} and \ref{lem:projection_inf} follows the sprit of projection on to the $\ell_1$ norm cone and $\ell_\infty$ cone, which can be considered as a special case with $\zeta=1$. The proof mainly applies fundamental results such as KKT condition in optimization. We include their proofs for completeness. 

\noindent{\bf Proof to Lemma~\ref{lem:projl1cone}}
\begin{proof}
Let us consider a trivial case first: $\|\a\|_1 \leq b$. In this case, the optimal values for $\x$ and $y$ are $\a$ and $b$ respectively. 

Then let us consider the nontrivial case: $\|\a\|_1 > b$. In this case, it is easy to see that the optimal values $\x^*$ and $y^*$ satisfy $\|\x^*\|_1 = y^*$ and the element signs of $\x^*$ are the same as $\a$. Therefore, to simplify the following notation and discussion, we make an assumption without the loss of generality
\[
\a_1\geq \a_2\geq \cdots \geq \a_d \geq 0.
\]
Based on this assumption, the problem \eqref{eq:proj1cone} is equivalent to solving the following problem
\begin{equation}
\begin{aligned}
\min_{\x, y}\quad & {1\over 2} \|\x- \a\|^2 + {\zeta \over 2}(y-b)^2
\\ \text{s.t.}\quad & \1^\top \x = y
\\& \x \geq {\bf 0}.
\end{aligned}
\end{equation}
It can be further simplified by
\begin{equation}
\begin{aligned}
\min_{\x}\quad & {1\over 2} \|\x- \a\|^2 + {\zeta \over 2}(\1^{\top}\x-b)^2
\\\text{s.t.}\quad & \x \geq \0.
\end{aligned}
\end{equation}
The optimal solution is defined by the KKT condition
\[
0\leq \x_i~\bot~\x_i - \a_i + \zeta(\1^\top \x - b) \geq 0\quad \forall i \in \{1,2,\cdots, d\}.  
\]
We also note that if $\a_i \geq \a_j$, then $\x^*_i \geq \x^*_j$ (otherwise we can simply swap the values of $\x^*_i$ and $\x^*_j$ to obtain a lower objective value). Therefore, from the monotonicity assumption on $\a$, we have
\[
\x^*_1 \geq \x^*_2 \geq \cdots \geq \x^*_d. 
\]
Let $j\in \{1,2, \cdots, d\}$ be the watershed: $\x^*_i = 0$ for $i>j$ and $\x^*_i \geq 0$ for $i\leq j$. The KKT is simplified by
\begin{align*}
\forall i\leq j\quad& \x_i = \a_i-\zeta(\1^\top \x - b) \geq 0 \\
\forall i>j\quad & \x_i=0\quad \a_i-\zeta(\1^\top \x - b) \leq 0.
\end{align*}  
Summarize all $\x^*_i$'s to obtain
\begin{align*}
& \1^\top \x = \left(\sum_{i=1}^j \a_i\right) - j\zeta(\1^\top \x - b) \\
\Rightarrow & \1^\top \x = \frac{\left(\sum_{i=1}^j \a_i\right) + j\zeta b}{1+j\zeta} \\
\Rightarrow & \zeta(\1^\top \x - b) = \frac{\left(\sum_{i=1}^j \a_i\right) - b}{\zeta^{-1} + j} =: \delta_j.
\end{align*}
Then finding a point satisfying the KKT condition is equivalent to finding a $j$ such that
\begin{equation}
\begin{aligned}
\forall i\leq j\quad& \x_i = \a_i-\delta_j \geq 0 \\
\forall i>j\quad & \x_i=0\quad \a_i-\delta_j \leq 0.
\end{aligned}
\label{eq:lem:proof_2} 
\end{equation}
Due to the monotonicity, to find such ``$j$'', we only need to enumerate all possible values for $j$ such that
\begin{equation}
\begin{aligned}
\a_j-\delta_j & \geq 0 \\
\a_{j+1}-\delta_j & \leq 0.
\end{aligned}
\label{eq:lem:proof_3}
\end{equation} 
As long as we find such ``$j^*$'', we can compute the optimal values for $\x^*$ from \eqref{eq:lem:proof_2} 
\[
\x^* = \max(\0, \a - \delta_{j^*})
\]
and $y^*$ from $y^* = \1^\top \x^*$. If we remove the assumption, the definition of $\delta_j$ should be modified into
\[
\delta_j:=\frac{\left(\sum_{i=1}^j |\a|^{(i)}\right) - b}{\zeta^{-1} + j},
\]
where $|\a|^{(i)}$ denotes the $i$th largest absolute value in $\a$. The condition to define the optimal $j$ in \eqref{eq:lem:proof_3} should be replaced by
\begin{equation*}
\begin{aligned}
|\a|^{(j)}-\delta_j & \geq 0 \\
|\a|^{(j+1)}-\delta_j & \leq 0.
\end{aligned}
\end{equation*}
The optimal values for $\x^*$ and $y^*$ are given by respectively
\[
\x^* = \text{sgn}(\a) \odot \max(\0, |\a|-\delta_{j^*})\quad y^*=\|\x^*\|_1.
\]
Algorithm~\ref{alg:projl1cone} exactly follows the procedure to find the optimal solution. It completes the proof.\footnote{We derive this solution from the perspective of the L one norm cone projection. The solution to \eqref{eq:lem:proj:l1} can also be derived from the perspective of shrinkage operator \citep{kowalski2013social}.}
\end{proof}

\noindent{\bf Proof to Lemma~\ref{lem:projection_inf}}
\begin{proof}
First we have that \eqref{eq:lem_proj} is equivalent to 
\begin{equation}
\min_{y\geq 0} ~ \min_{\|\x\|_\infty \leq y} 
{1\over 2} \left(\|\x-\a\|^2 + \zeta(y-b)^2\right).
\label{eq:lem:proof:1}
\end{equation}
Given $y$, the optimal value for $\x$ is 
\begin{equation}
\x_i = \text{sgn}(\a_i)\odot \min (|\a_i|, y) \quad \forall i
\label{eq:lem:proof:0}
\end{equation}
Replacing $\x$ in \eqref{eq:lem:proof:1} by its optimal value with respect to $\a$, we obtain
\begin{equation}
\min_{y\geq 0} {\zeta\over 2} (y-b)^2 + {1\over 2} \sum_{i=1}^d\min(y-|\a_i|,0)^2.
\end{equation}
The optimality condition is
\[
0\leq \zeta(y-b) + \sum_{i=1}^d\min(y - |\a_i|, 0) \perp y \geq 0.
\]
To decide the optimal value for $y$, we can first test if $y=0$ is the optimal solution. $y=0$ is the optimal solution is equivalent to $\zeta(0-b) + \sum_{i=1}^d\min(0 - |\a_i|, 0) \geq 0$. If $y=0$ is not the optimal solution, we know that the optimal solution must be greater than $0$. Thus, we only need to find a $y>0$ satisfying
\[
0 = \zeta(y-b) + \sum_{i=1}^d\min(y - |\a_i|, 0).
\]
We do not have a closed form solution to this equality due to the $\min(\cdot, \cdot)$ function. To remove this part, we sort $\{|\a_i|~i=1,\cdots d\}$ and consider different ranges for $y$:
\[
(0,|\a|^{(1)}],~(|\a|^{(1)}, |\a|^{(2)}],~\cdots,~ (|\a|^{(d-1)},~|\a|^{(d)}],~(|\a|^{(d)},~\infty)
\] 
where $|\a|^{(i)}$ denotes the $i$th smallest element of $\a$ in the absolute value sense. For the range $(|\a|^{(d)},~\infty)$, we only need to check if
\[
y = b \in (|\a|^{(d)},~\infty)
\] 
If yes, we find the optimal solution for $y^*$. Otherwise, we check the remaining ranges. For the range $(0,|\a|^{(1)}]$, we only need to check if
\[
y={1\over \zeta+d} \left(\zeta b+ \sum_{i=1}^d |\a|^{(i)}\right) \in (0, |\a|^{(1)}].
\]
If yes, we find the optimal solution for $y^*$. For the general range $(|\a|^{(k)},~|\a|^{(k+1)}]$, we only need to check if 
\[
y={1\over \zeta+d-k} \left(\zeta b+ \sum_{i=k+1}^d |\a|^{(i)}\right) \in (|\a|^{(k)},~|\a|^{(k+1)}].
\]
When the optimal value for $y^*$ is determined, one easily recover the optimal value for $\x^*$ by \eqref{eq:lem:proof:0}. Algorithm~\ref{alg:projection_inf} exactly follows the procedure above. It completes the proof.
\end{proof}

{\noindent \bf Proofs to Theorem~\ref{thm:esvm-re}}
\begin{proof}
Denote $y_iX_i$ by $Z_i$($i\in \{1,2,\cdots, t\}$). To deal with the first non-smooth term ``$\max(1-y_iX_i^{\top}\w, 0)$'', we rewrite \eqref{eq:ESVM} by introducing a vector $\bfXi$ in the following:
\begin{equation}
\begin{aligned}
\min_{\w, \bfXi \geq 0} \quad &  \1^{\top} \bfXi + {\alpha \over 2}\|\w\|^2 +  {\beta \over 2} \|\w\|_e^2
\quad  \quad
\\ \text{s.t.} \quad &
1 - Z_i^{\top}\w \leq \bfXi_i\quad & \forall i 
\end{aligned}
\label{eq:ESVM_1}
\end{equation}
We introduce a dual variable $\u \in \R^t$ to move the constraint ``$1 - Z_i^{\top}\w \leq \bfXi_i$'' to the objective and apply Lemma~\ref{lem:dual} to rewrite the exclusive sparsity regularizer:
\begin{equation*}
\begin{aligned}
\min_{\w,~\bfXi\geq 0} \max_{\u\geq 0, \v^g\in\R^{|g|}, g \in \mathcal{G}} \quad &  \1^{\top} \bfXi + {\alpha \over 2}\|\w\|^2 + \langle \u, \1-Z^{\top}\w-\bfXi \rangle + \\ &  \sum_{g\in \mathcal{G}}\langle \v^g, \w_g\rangle - {1 \over 2\beta}\|\v^g\|_\infty^2
\end{aligned}
\end{equation*}	
which is equivalent to \eqref{eq:ESVM_1}. 
From the definition of $\overline{\v_g}$, we have $\sum_{g\in \mathcal{G}} \langle \v^g, \w_g \rangle = \left\langle \sum_{g\in \mathcal{G}} \overline{\v^g}, \w \right\rangle$.

From strong duality, we safely swap $\min$ and $\max$:
{\small
\begin{align}
\nonumber
&\max_{\u\geq 0, \v^g\in\R^{|g|}, g \in \mathcal{G}} \min_{\w,~\bfXi\geq 0}   \1^{\top} \bfXi + {\alpha \over 2}\|\w\|^2 + \langle \u, \1-Z^{\top}\w-\bfXi \rangle + \\
\nonumber & \quad\quad\quad\quad\quad\quad\quad\quad\quad\quad \left\langle \sum_{g\in \mathcal{G}} \overline{\v^g}, \w \right\rangle - {1 \over 2\beta}\sum_{g\in \mathcal{G}}\|\v^g\|_\infty^2
\nonumber \\ = &
\max_{\u\geq 0, \v^g\in\R^{|g|}, g \in \mathcal{G}}\min_{\bfXi\geq 0} \quad 
- {1\over 2\alpha} \left\|Z \u - \sum_{g\in \mathcal{G}}\overline{\v^g}\right\|^2 
+ \langle \bfXi,1-\u \rangle + 
\nonumber \\ & \quad\quad\quad\quad\quad\quad\quad\quad\quad\quad \1^{\top} \u  - {1 \over 2\beta}\sum_{g\in \mathcal{G}}\|\v^g\|_\infty^2
\nonumber  \\ = & 
\max_{\u\geq 0, \v^g\in\R^{|g|}, g \in \mathcal{G}}~ \begin{cases}
  - {1\over 2\alpha} \left\|Z \u - \sum_{g\in \mathcal{G}}\overline{\v^g}\right\|^2+ \\ \1^{\top} \u - {1 \over 2\beta}\sum_{g\in \mathcal{G}}\|\v^g\|_\infty^2,& \text{if } \u\leq 1\\
    -\infty,              & \text{otherwise}
\end{cases}
\label{eq:ESVM_2}
\end{align}	
}
where the second equality is obtained by letting 
\begin{equation}
\quad \w={1\over \alpha}\left(Z \u - \sum_{g\in \mathcal{G}}\overline{\v^g}\right).
\label{eq:ESVM_3}
\end{equation}	
Eq.~\eqref{eq:ESVM_2} essentially defines the dual problem of \eqref{eq:ESVM}
\begin{align}
\min_{\u, \v^g~g\in \mathcal{G}}\quad &  \underbrace{{1\over 2\alpha} \left\|Z \u - \sum_{g\in \mathcal{G}}\overline{\v^g}\right\|^2 - \1^{\top} \u }_{=:F([\u; \{\v^g\}_{g\in \mathcal{G}}])} + 
\nonumber \\ & \underbrace{{1 \over 2\beta}\sum_{g\in \mathcal{G}}\|\v^g\|_\infty^2 + \I_{\0\leq \u \leq \1}(\u)}_{=:H([\u; \{\v^g\}_{g\in \mathcal{G}}])}. 
\end{align}
It completes the proof.
\end{proof}

\noindent{\bf Proof to Theorem~\ref{thm:even}}
\begin{proof}
We use $\Omega$ to denote the index set of useful features and $\Omega_j \subseteq g_j$ to denote the index set of useful features assigned to group $j$, where $j\in \{1,2,\cdots, m\}$. In most cases, the Uniform type model is difficult to analyze due to the dependence among variables. A common trick is to use the Bernoulli model to approximate it and then bridge the difference between the Uniform model and Bernoulli model. The Bernoulli model performs the following experiment: for any group $g_j$, each index in $g_j$ corresponds to a useful feature with probability $p$ and a useless feature with probability $(1-p)$. The Bernoulli model admits the independence among all indices, thus is much easier to derive the probability bound. 
Denote by $\PU$ and $\PB$ probabilities calculated under the uniform and Bernoulli models. For example, $\P_{\text{Unif}(s)}\left(\max_{j\in\{1,2,\cdots, m\}}|\Omega_j| \leq 2{s\over m}\right)$ is the probability of the number of useful features assigned into any group is less than $2s/m$ by using the uniformly randomly assignment.
\begin{align*}
& \PB\left(\max_j |\Omega_j|< {s\over m} (1+t)\right) 
\\ = & 
\sum_{k=0}^n \{\PB\left(\max_j |\Omega_j|< {s\over m} (1+t)~\Bigg|~|\Omega|=k\right) 
\\ &
\quad\quad \PB(|\Omega|=k)\}
\\ = & 
\sum_{k=0}^{s-1} \{\PB\left(\max_j |\Omega_j|< {s\over m} (1+t)~\Bigg|~|\Omega|=k\right) 
\\ &
\quad\quad \PB(|\Omega|=k)\}~+
\\ & 
\sum_{k=s}^n \{\PB\left(\max_j |\Omega_j|< {s\over m} (1+t)~\Bigg|~|\Omega|=k\right) 
\\ &
\quad\quad \PB(|\Omega|=k)\}
\\ \leq &
\PB(|\Omega| < s)~+
\\ & \sum_{k=s}^n \{\PB\left(\max_j |\Omega_j|< {s\over m} (1+t)~\Bigg|~|\Omega|=k\right) 
\\ &
\quad\quad \quad \PB(|\Omega|=k)\}
\\ = &
\PB(|\Omega| < s)~+
\\ & \sum_{k=s}^n \{\PU\left(\max_j |\Omega_j|< {s\over m} (1+t)\right) \PB(|\Omega|=k)\}
\\ \leq & 
\PB(|\Omega| < s)~+~\P_{\text{Unif}(s)}\left(\max_j |\Omega_j|< {s\over m} (1+t)\right)
\end{align*}
where the last equality uses the fact 
\begin{align*}
& \PB\left(\max_j |\Omega_j|< {s\over m} (1+t)~\Bigg|~|\Omega|=k\right) \\ = & \PU\left(\max_j |\Omega_j|< {s\over m} (1+t)\right)
\end{align*}
and the last inequality uses the monotonicity
\begin{align*}
& \P_{\text{Unif}(s)}\left(\max_j |\Omega_j|< {s\over m} (1+t)\right) \\ \geq & \P_{\text{Unif}(s+1)}\left(\max_j |\Omega_j|< {s\over m} (1+t)\right).
\end{align*}
It follows that
\begin{align}
\nonumber & \P_{\text{Unif}(s)}\left(\max_j |\Omega_j|\geq {s\over m} (1+t)\right) \\ \leq &  \PB\left(\max_j |\Omega_j|\geq {s\over m} (1+t)\right) + \PB(|\Omega| < s).
\label{eq:proof_1} 
\end{align}


Bernstein's inequality states that for a sum of uniformly bounded independent random variables with $Y_i - \E Y_i \le c$, we have
\[
\P \left(\sum_{i=1}^k (Y_i - \E Y_i) \geq \delta \right) \leq \exp \left\{-{\delta^2 \over 2k\sigma^2 + 2c\delta/3}\right\}\quad 
\] 
and
\begin{align}
\label{eq:Bern}
\P \left(\sum_{i=1}^k (Y_i - \E Y_i) \leq -\delta \right) \leq \exp \left\{-{\delta^2 \over 2k\sigma^2 + 2c\delta/3}\right\}
\end{align}
hold for any $\delta>0$, where $\text{Var}(Y_i) = \sigma^2$. Let $Y_i$'s follow i.i.d. Bernoulli distribution with parameter $p$. 
Take $p={s\over n}(1+\epsilon)$ where $t>\epsilon > 0$. From $\E (Y_i) = p$,  $c = 1 - p$ and $\text{Var} (Y_i) = (1-p)p$ for $i\in \{1,\cdots, n/m\}$, we have
\begin{align*}
& \PB\left(\max_j |\Omega_j| \geq {s \over m} (1+t) \right) 
\\ \leq & \sum_j \PB\left(|\Omega_j| \geq {s \over m} (1+t) \right) 
\\ = & \sum_j \PB\left(|\Omega_j| \geq {s \over m}(1+\epsilon) + {s\over m}(t-\epsilon)\right)
\\
= & \sum_j \P \left(\sum_{i=1}^{n/m} (Y_i - \E Y_i) \geq {s\over m}(t-\epsilon)\right) \\
\leq & \sum_j \exp \left\{-{(t-\epsilon)^2 s^2 / m^2 \over 2(1-p)pn/m + 2(1-p)(t-\epsilon)s/(3m)} \right\}
\\
\leq & \exp \left\{-{(t-\epsilon)^2 s / m \over 2(1+\epsilon) + 2(t-\epsilon)/3} + \log m\right\}.
\end{align*}
Taking $\epsilon = 0.1 t \leq 1$, we have 
\begin{align}
\nonumber & \PB\left(\max_j |\Omega_j| \geq {s \over m} (1+t) \right) \\ \leq & \exp \left\{-0.08t^2 s / m+ \log m\right\}.
\label{eq:proof_2} 
\end{align}   
From Bernstein's inequality \eqref{eq:Bern}, we can similarly have
\begin{align}
\nonumber \PB(|\Omega| < s) & \leq \exp\{-\epsilon^2  s/ 2(1+{3\over 4} \epsilon)\} 
\\ & \leq \exp\{-0.001t^2 s\}.
\label{eq:proof_3}
\end{align}
Plugging \eqref{eq:proof_2} and \eqref{eq:proof_3} into \eqref{eq:proof_1}, we obtain
\begin{align}
\nonumber & \P_{\text{Unif}(s)}\left(\max_j |\Omega_j| \geq {s\over m}(1+t)\right) 
\\ \leq & \exp\left\{-0.001t^2 s\right\} + \exp\left\{-0.08{t^2 s\over m} + \log m\right\}
\label{eq:proof_max}
\end{align}

Next we turn to prove $\P_{\text{Unif}(s)}\left(\min_j |\Omega_j| \leq {s\over m}(1-t)\right)$ holds with high probability. Choose $p= {s\over m}(1-\epsilon)$ where $0 < \epsilon = 0.1 t $.
\begin{align}
\nonumber
& \PB \left(\min_j |\Omega_j| \leq {s\over m}(1-t)\right) 
\\
\nonumber \geq & \sum_{k=0}^s \PB\left(\min_j |\Omega_j| \leq {s\over m}(1-t)~\Bigg|~|\Omega| = k\right) \PB(|\Omega| = k) 
\\
\nonumber
= & \sum_{k=0}^s \P_{\text{Unif}(k)}\left(\min_j |\Omega_j| \leq {s\over m}(1-t)\right) \PB(|\Omega| = k)
\\
\nonumber
\geq & \P_{\text{Unif}(s)}\left(\min_j |\Omega_j| \leq {s\over m}(1-t)\right) \PB(|\Omega| \leq s).
\end{align}
If follows that
\begin{align}
 \nonumber 
 & \P_{\text{Unif}(s)}\left(\min_j |\Omega_j| \leq {s\over m}(1-t)\right) 
 \\
 \leq & \frac{\PB \left(\min_j |\Omega_j| \leq {s\over m}(1-t)\right)} {\PB(|\Omega| \leq s)} 
\label{eq:proof_4}
\end{align}

Since $s \geq np$, we have
\begin{align}
\PB(|\Omega| \leq s) \geq 0.5.
\label{eq:proof_5}
\end{align}
From Bernstein's inequality, we can similarly have
\begin{align}
\PB\left(\min_j |\Omega_j| \leq {s \over m} (1-t) \right) \leq \exp \left\{-0.2t^2 s / m+ \log m\right\}.
\label{eq:proof_6}
\end{align}
Combine \eqref{eq:proof_4}, \eqref{eq:proof_5}, and \eqref{eq:proof_6} to obtain
\begin{align}
 \P_{\text{Unif}(s)}\left(\min_j |\Omega_j| \leq {s\over m}(1-t)\right) \leq 2 \exp \left\{-0.08t^2 s / m+ \log m\right\}.
 \label{eq:proof_min}
\end{align}

We can prove the final result by combining \eqref{eq:proof_max} and \eqref{eq:proof_min}:
\begin{align*}
& \P_{\text{Unif}(s)} \left(\frac{\max_j |\Omega_j|}{\min_j |\Omega_j|} \geq \frac{1+t} {1-t} \right) 
\\
\leq & \P_{\text{Unif}(s)}\left(\min_j |\Omega_j| \leq {s\over m}(1-t)\right) + 
\\
&\P_{\text{Unif}(s)}\left(\max_j |\Omega_j| \geq {s\over m}(1+t)\right)
\\
\leq & 3\exp\left\{-0.08t^2 s / m+ \log m\right\} + \exp\left\{-0.001t^2 s\right\}
\end{align*}

If we have $s=O(t^{-2}m\log m)$, the probability in the right hand side of \eqref{eq:proof_4} converges to zero, which indicates that the probability on the left hand side converges to $1$. It completes the proof.
\end{proof}

\end{document}